\documentclass[letterpaper]{article} 
\usepackage{aaai24}  
\usepackage{times}  
\usepackage{helvet}  
\usepackage{courier}  
\usepackage[hyphens]{url}  
\usepackage{graphicx} 
\def\UrlFont{\rm}  
\usepackage{natbib}  
\usepackage{caption} 
\usepackage{algorithm}
\usepackage{algorithmic}

\usepackage{newfloat}
\usepackage{listings}
\usepackage{xcolor}
\DeclareCaptionStyle{ruled}{labelfont=normalfont,labelsep=colon,strut=off} 
\floatstyle{ruled}
\newfloat{listing}{tb}{lst}{}
\floatname{listing}{Listing}
\usepackage{xcolor}

\title{MixNet: Toward Accurate Detection of Challenging Scene Text in the Wild}
\author{
    Yu-Xiang Zeng,
    Jun-Wei Hsieh,
    Xin Li,
    Ming-Ching Chang,
}
\affiliations{
    leo32656884602@gmail.com,
    jwhsieh@nycu.edu.tw, 
    Xin.Li@mail.wvu.edu,
    mchang2@albany.edu
}

\usepackage{bibentry}

\usepackage{multirow}

\begin{document}

\maketitle

\begin{abstract}
Detecting small scene text instances in the wild is particularly challenging, where the influence of irregular positions and nonideal lighting often leads to detection errors. We present MixNet, a hybrid architecture that combines the strengths of CNNs and Transformers, capable of accurately detecting small text from challenging natural scenes, regardless of the orientations, styles, and lighting conditions. MixNet incorporates two key modules: (1) the Feature Shuffle Network (FSNet) to serve as the backbone and (2) the Central Transformer Block (CTBlock) to exploit the 1D manifold constraint of the scene text. We first introduce a novel {\bf feature shuffling} strategy in FSNet to facilitate the exchange of features across multiple scales, generating high-resolution features superior to popular ResNet and HRNet. The FSNet backbone has achieved significant improvements over many existing text detection methods, including PAN, DB, and FAST. Then we design a complementary CTBlock to leverage {\bf center line} based features similar to the medial axis of text regions and show that it can outperform contour-based approaches in challenging cases when small scene texts appear closely. Extensive experimental results show that MixNet, which mixes FSNet with CTBlock, achieves state-of-the-art results on multiple scene text detection datasets.
\end{abstract}


\begin{figure*}[t]
\centerline{
  \includegraphics[width=\linewidth]{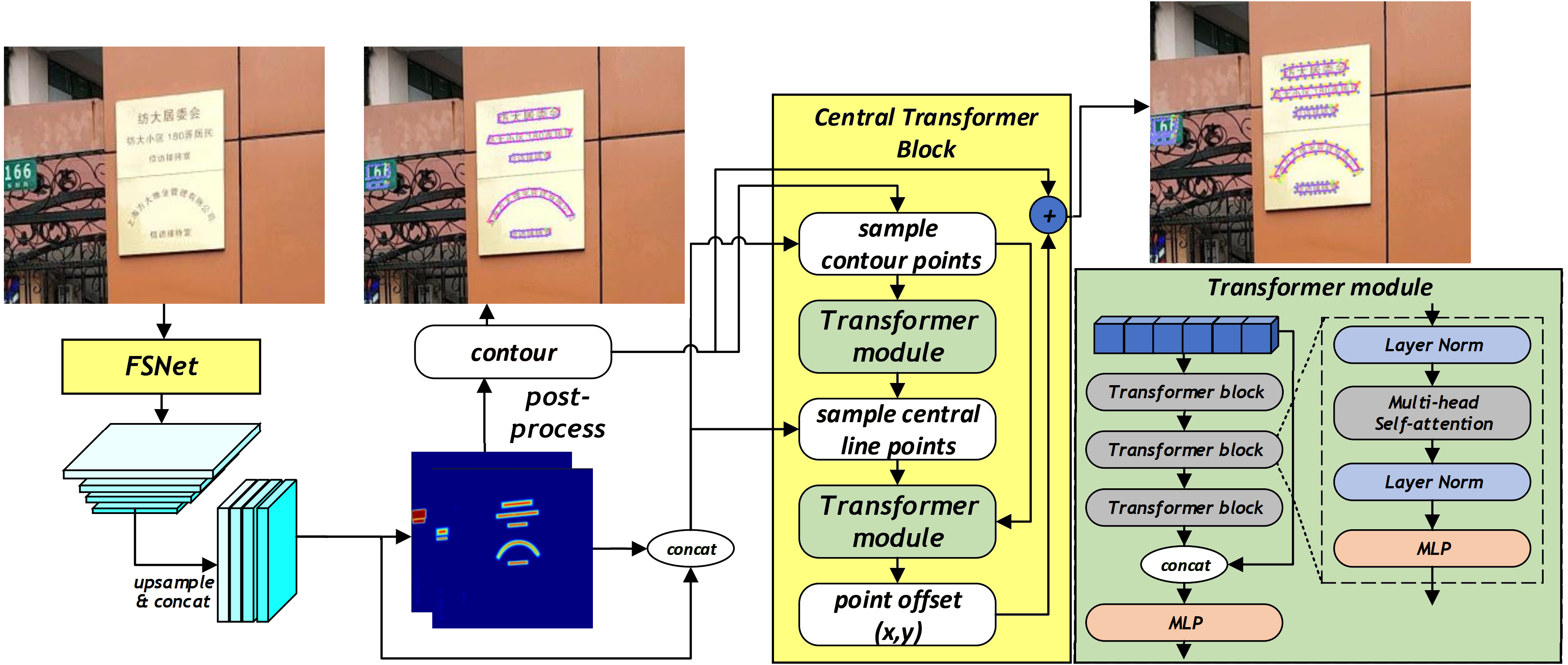}
  \vspace{-2mm}
}
\caption{MixNet is a hybrid architecture for scene text detection that includes two novel components: (1) the {\bf Feature Shuffling Network (FSNet)} for generating high-resolution features and (2) the {\bf Central Transformer Block (CTBlock)} for detecting the medial axis of text regions (called ``center line'' in this paper). MixNet (FSNet+CTBlock) is particularly effective in detecting small and curved text in natural scenes, {\em e.g.}, fine prints on the plaque.}
\label{fig:overview}
\vspace{-4mm}
\end{figure*}

\section{Introduction}
\label{introduction}

Scene text detection is a common task in computer vision with various practical applications in scene understanding, autonomous driving, real-time translation, and optical character recognition. Deep learning-based methods continuously set new performance records on scene text detection benchmarks. In natural scenes, text boxes can appear in arbitrary orientations and non-rectangular shapes. Segmentation-based methods are widely used to handle texts of arbitrary shapes. Unlike object detection methods that rely on bounding boxes, segmentation-based methods predict pixel-level masks to identify texts in each area. However, these methods often rely on Convolutional Neural Networks (CNNs), which tend to ignore the global geometric distribution of the overall text boundary layout, leading to the following two problems. First, CNN focuses on local spatial features and thus is sensitive to noise in text regions. Second, commonly used CNN backbones, such as ResNet~\cite{resnet} and VGG~\cite{vgg}, provide rough high-resolution features, which are useful for large text detection but not conducive to the detection of small text instances.

Transformer~\cite{transformer} has gained remarkable success in multiple fields and provides an alternative method to extract features. Unlike CNN, which focuses on local features of adjacent regions, Transformer emphasizes the global spatial relationships between text regions. Along with extensive research on Transformer, several contour-based methods have recently been developed that directly detect text contours and achieve state-of-the-art performance. For example, TESTR~\cite{testr} used the Transformer encoder to extract a larger range of text features and send the Region of Interest (ROI) to the decoder to generate a text box based on the box-to-polygon flow. DPText-DETR~\cite{dptext} performed point sampling in text ROI to obtain better location information. These contour-based methods can generate text contours directly from input images, thus eliminating the need for post-processing in segmentation-based methods.

In this paper, we propose a hybrid architecture named MixNet to combine the strengths of CNNs and Transformers (see Fig.~\ref{fig:overview}). We hypothesize that changing the backbone architecture is necessary to provide powerful high-resolution features and reduce the interferences {\em e.g.}, noise contamination and illumination variation that hinders accurate detection, especially in the presence of small and curved texts. We have empirically observed that low-resolution features are less affected by noise. Therefore, it is desirable to develop a scale-invariant backbone for feature extraction. We propose a novel {\bf Feature Shuffle Network (FSNet)} to exchange features between low-resolution and high-resolution layers during feature extraction, which deepens the extraction layer of high-resolution features. FSNet allows the backbone to generate better features than other popular backbones such as ResNet and HRNet~\cite{hrnet}. 

After extracting features, we propose a new {\bf Central Transformer Block (CTBlock)} to further enhance the global geometric distribution of the text regions. Specifically, the heatmap generated by the image features is post-processed to obtain a rough contour of the text. Based on this rough text contour, we sample image features that represent positional information and features of the text contour. These sampled features are then used as input for the Transformer module to learn the geometric distribution of text boundaries and generate the center lines of text contours. Traditional contour-based methods cannot detect two adjacent text lines well if they are too close, which makes their contours on the heatmap interfere with each other. However, their center lines are still separated. Thus, we sample the feature points along their respective center lines and merge the contour feature together so that the above occlusion problem can be well addressed. The merged features are fed into the next Transformer module to calculate the vertex offset for each point. A precise text contour is then generated by applying the vertex offsets to the initial contour of the text.

The contribution of this paper includes:
\begin{itemize}
\item We designed a new backbone called FSNet to extract features while exchanging features between low- and high-resolution layers. This backbone produces better image features and reduces noise interference in high-resolution feature representations. Additionally, replacing the backbone of other methods with FSNet can unanimously improve detection performance.

\item We propose a method called Central Transformer Block (CTBlock) that mixes point-sampled features of contours and center lines. It helps the model learn and refine text boundaries, resulting in more precise text contours.

\item The proposed MixNet (FSNet+CTBlock) sets a new state-of-the-art for representative arbitrary-shape scene text detection benchmarks. Improvements are reported for three popular datasets of scene text detection. MixNet is particularly effective at detecting small curved text in natural scenes.
\end{itemize}

\begin{figure*}[t]
\centerline{
  \includegraphics[width=\linewidth]{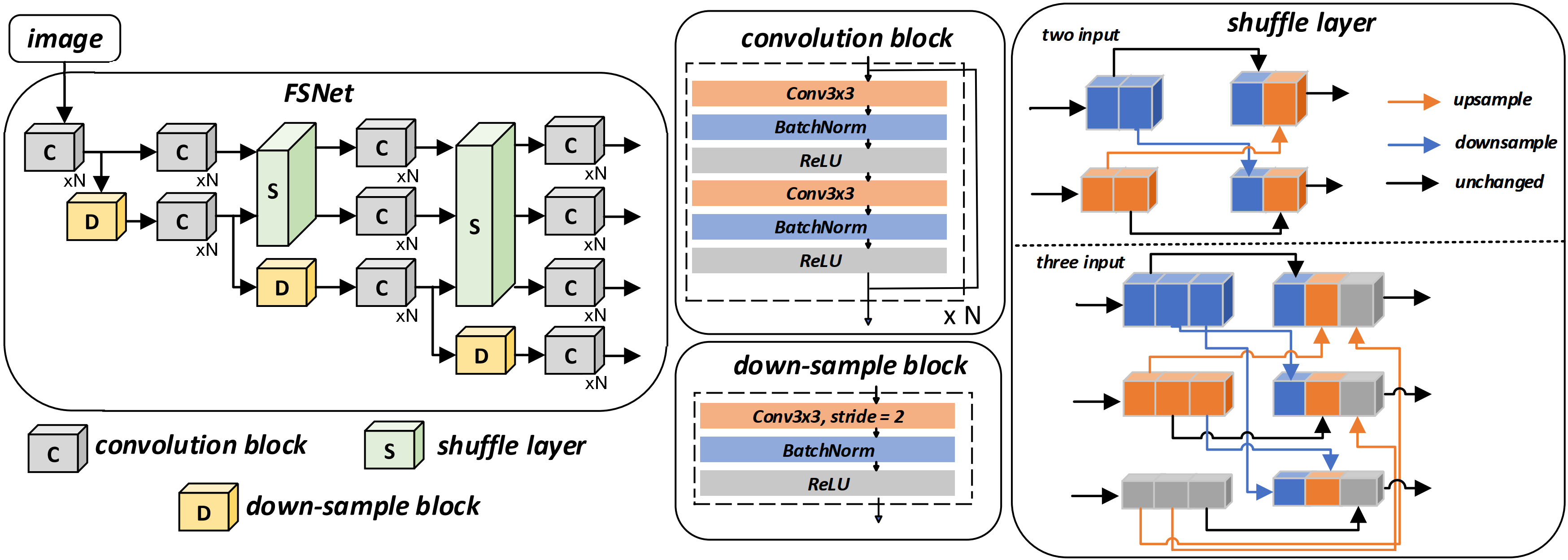}
  \vspace{-2mm}
}
\caption{Detailed architecture of FSNet: feature shuffle is implemented by two shuffle layers. Detailed implementations of the three modules ({\bf C, D, S}) are shown on the right. Note that FSNet contains two shuffle layers. The first shuffle layer has two-resolution feature inputs (top-right) and the second shuffle layer has three-resolution feature inputs (bottom-right).
}
\label{fig:FSNet}
\vspace{-4mm}
\end{figure*}

\section{Related works}
\label{related works}

\subsection{Segmentation-based Scene Text Detection}

Segmentation-based scene text detection model generates a heatmap indicating the likelihood of each image pixel being a text instance. The heatmap is then processed to produce several text contours. This approach faces a major challenge: if the predicted heatmap connects multiple text instances, the post-processing algorithms cannot separate them correctly. To address this issue, most segmentation-based methods predict the text kernel rather than the text instance, reducing the likelihood of text instances being connected by reducing their size. DB~\cite{DB} proposes an additional method of predicting the threshold map to require higher confidence values for the border of the text instance to deal with the problem of connecting text instances. On the other hand, PAN~\cite{PAN} predicts the embedding value of each pixel, with each text instance having a different embedding value. Pixel aggregation post-processing is used to connect pixels with the same embedding value. By differentiating the embedding value, multiple text instances can be separated into multiple text contours by post-processing, even if they are connected.

Segmentation-based methods has a speed issue when running the post-processing algorithm on CPU to convert the predicted heatmap into text contours. Therefore, methods such as DB and PAN use a lightweight backbone to achieve real-time scene text detection. FAST~\cite{fast} proposes a lightweight network constructed by Neural Architecture Search (NAS) with a post-processing algorithm that uses the GPU to perform part of the steps. The inference time of FAST is much faster due to the lightweight architecture and optimized post-processing. However, their capability of handing small or curved texts remains questionable. 

\subsection{Contour-based Scene Text Detection}

Contour-based scene text detection methods utilize mathematical curves to fit the text of arbitrary shapes. The model predicts the parameters of the curve that fit the text. As the text shape becomes more complex, the model requires more parameters, and the fitting becomes more prone to failure. However, contour-based methods have an end-to-end structure for generating text boxes, directly outputting the control points of the curve to represent the text outline. This eliminates the need for post-processing required by segmentation-based methods. Additionally, contour-based methods can generate overlapping text boxes, which cannot be achieved by segmentation-based methods.

ABCNet~\cite{abcnet} is the first architecture to predict control points of Bezier curves that fit the text of arbitrary shapes. TextBPN~\cite{textBPN} uses dilated convolution to generate rough boundary and offset information and then iteratively deforms the boundary to generate more accurate text boundaries. FCENet~\cite{fce} enhances the ability to predict highly curved texts by introducing Fourier contour embedding. Recently, inspired by Transformer, FSG~\cite{fsg} has used a Transformer encoder to predict more accurate curve parameters. TESTR designs a box-to-polygon process that uses anchor box proposals from the Transformer encoder to generate positional queries, which facilitates polygon detection. 
However, such a bounding-box representation might lack precision. DPText-DETR~\cite{dptext} conducts point sampling directly in the text bounding box to address this issue of TESTR.
In our study, we use heatmaps to extract both contours and center lines (approximating the medial axis of text regions) from each text instance. 

\section{The Proposed Method}
\label{methods}

\subsection{Overview of MixNet}

Fig.~\ref{fig:overview} illustrates the overall architecture of MixNet. FSNet is used as the backbone network to extract text features and generate pixel-level classification, distance field, orientation field, and embedding values. This information is then used to generate a rough text contour. The $P$ points sampled from the rough text contour are taken as input to the CTBlock. Specifically, the $P$ sample points and their corresponding image features are entered into the CTBlock as a sequence. Then, the first Transformer module predicts $x$ and $y$ offsets of the sample points and uses them to generate the center line of a text contour. In the same way, the center line is sampled and combined with the feature sequence of the rough contour and sent to the second Transformer module to correct the rough text contour and produce a finer text contour. In this section, we first explain in detail the backbone of FSNet and then the process of predicting point offsets using the CTBlock.

\subsection{Feature Shuffle Network (FSNet)}

CNN-based backbones can be sensitive to noise during feature extraction. Popular backbones such as ResNet and VGG often produce rough high-resolution features that are not well suited to detect small text instances. To address these challenges, we propose FSNet, which incorporates a network that can better extract high-resolution features. Additionally, we empirically observed that low-resolution features have better ability against noise and perturbations. Thus, FSNet is designed to exchange both low- and high-resolution features during feature extraction, making the extracted high-resolution features less vulnerable to noise. These two characteristics improve the FSNet backbone's capability for detecting small-sized texts. The architecture of FSNet is similar to HRNet~\cite{hrnet}, but with a key difference in the way FSNet {\em shuffles features}. While HRNet mixes features between layers by adding them, FSNet equally divides the channels of each resolution and shuffles the divided features. After shuffling, the cut features of each resolution are up-sampled and down-sampled to the same size and concatenated into new features.

FSNet contains three main modules, namely the {\em convolution block}, {\em down-sample block}, and {\em shuffle layer}, as shown in Fig.~\ref{fig:FSNet}. Convolution blocks are stacked in large numbers to extract features. The down-sample block uses a $3\times3$ convolution with a stride of two for down-sampling. Note that the stacking amount of each convolution block is different. FSNet contains two shuffle layers (green in Fig.~\ref{fig:FSNet}). The first shuffle layer has two-resolution feature inputs and the second shuffle layer has three-resolution feature inputs. The shuffle layer divides the channels of the features of each resolution into the number of inputs $N$. The input feature $F^i$ is divided into $N$ cut features $F^i_1$,...,$F^i_n$,...,$F^i_N $, where $F^i_n$ denotes the $n$-${th}$ input at the $i$-${th}$ scale. The cropped features are up-sampled or down-sampled to the corresponding size according to the index and then concatenated to form new features. The shuffling operation can facilitate the exchange of features across multiple scales, generating more discriminative features superior to the popular ResNet and HRNet~\cite{hrnet}. Since the shuffle layer has no learnable parameters, it is more efficient than HRNet. 
At the final layer of FSNet, the results of all four scales are concatenated into a single feature map. 


\subsection{Central Transformer Block (CTBlock)}


In prior work, sampling points were located primarily in the peripheral vicinity of text instances. For example, points were sampled along the bounding box of each text region in DPText-DETR \cite{dptext}. However, such peripheral sampled features often encompass numerous background attributes, which prevents focusing on just the text. To overcome this challenge, we present {\bf Central Transformer Block (CTBlock)}, an innovative design that integrates the center line with the corresponding features to represent each text region. 
The Transformer module within CTBlock adopts an encoder-decoder structure. As shown in Fig.~\ref{fig:overview} right, a three-layer transformer block is included as the encoder. Each transformer block contains a multihead self-attention block and a Multi-Layer Perceptron (MLP) network. The decoder consists of a simple MLP.

The key advantage of the proposed transformer module is that the combination of peripheral and center line features can produce a more precise text contour. When two text instances are positioned closely, the contour features of the neighboring regions might lead to deformation or even interference. The center line effectively maintains the separation of these two text instances in such scenarios. In practice, a rough contour of each text instance is extracted based on the heatmap generated by the backbone network. Then we select $N$ points along each contour to represent the text contour. The length of each contour, denoted $C_i$, is $L_i$, where $i$ is the index of each text instance. We divide $L_i$ into $N$ equal parts of length $T$. A point is sampled every time $T$ and passed from the starting point. After repeating this process $N$ times, we obtain $N$ points along the text contour. We empirically observed a slight performance improvement with $N \geq 20$. To maintain a balance between inference time and performance, we set $N = 20$ in our experiments. The corresponding image features and heatmaps are extracted to form a sequence of features. Next, it passes through the first transformer module, which generates points that represent the center line. 

\begin{figure}[t]
\centerline{
  \includegraphics[width=\linewidth]{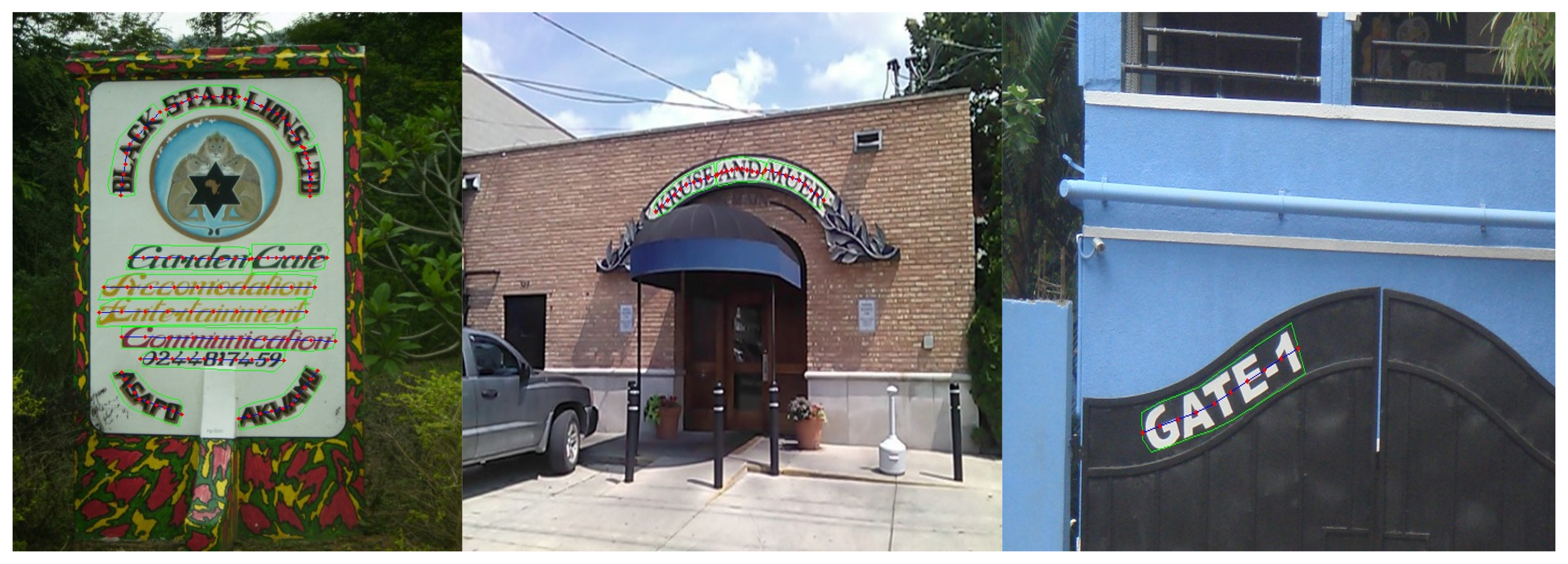}
  \vspace{-2mm}
}
\caption{The sample point of a {\bf center line} is shown as a red dot. We assume that scene text is positioned along either a straight line or a smooth curve in natural scenes ($i.e.$, a 1D manifold embedded into the 2D Euclidean space).}
\label{fig:centerline}
\vspace{-4mm}
\end{figure}

\begin{table*}[t]
\caption{Performance evaluation on both noise-free and noisy datasets. Note how FSNet demonstrates better noise-resistant performance over ResNet50.
\vspace{-3mm}
}
\label{table:noise}
\centerline{
\begin{tabular}{lccccccc}
\hline
Backbone	&Impulse noise &Prec.(\%)  &Recall(\%) &F1(\%) &Small(\%) &Medium(\%) &Large(\%)\\
\hline
ResNet50    & \multirow{2}{*}{0\%}	&91.8	&84.0	&87.7  &53.0 &88.0 &94.5\\
FSNet      &	&92.1	&87.7	&89.8 \textbf{(+2.1)} &68.8 &92.4 &97.6 \\
\hline
ResNet50	& \multirow{2}{*}{5\%}	&85.0	&68.3	&75.8  & 26.9 &68.3 &89.2\\
FSNet      &	&90.4	&72.1	&80.3 \textbf{(+4.5)} &42.3 &75.3 &92.0\\
\hline
ResNet50	&\multirow{2}{*}{10\%}	&83.4	&58.9	&69.1 &16.5 &57.2 &84.9\\
FSNet      &	&88.6	&65.2	&75.1 \textbf{(+6.0)}  &30.7 &63.8 &88.0\\
\hline
\end{tabular}
}
\end{table*}

Fig.~\ref{fig:centerline} illustrates some examples of such center line based scene text detection. During training, the center line points are supervised, and the ground truth for these points is obtained from the upper and lower contours of the text contour ground truth. As the number of ground truth points in each dataset may vary, we use the aforementioned sampling method to obtain a fixed number $C$ of center line points and $C$=10 in this paper. Similarly, after obtaining the center line points, we cropped the corresponding image features and heatmaps to form another feature sequence. Finally, by combining the feature sequence of text contour and the feature sequence of center line, the second Transformer module outputs the vertex offset for each point of the initial contour, resulting in a more precise and refined contour. 

\section{Experiments}
\label{experiments}

\begin{table}[t]
\caption{Performance evaluation on Total-Text dataset for three scene text detection methods with and w/o FSNet. FSNet unanimously improves precision, recall, and F1 scores for all three methods.
\vspace{-2mm}
}
\label{table:differentMethods}
\centerline{
\begin{tabular}{l|c|ccc}
\hline
Method	&FSNet	&Prec.(\%)	&Recall(\%)	&F1(\%) \\
\hline
\multirow{2}{*}{PAN}	&$\times$	&89.3	&81.0	&85.0 \\
                    &$\surd$ 	&89.5	&83.8	&86.5 \\
\hline
\multirow{2}{*}{DB}	&$\times$	&87.1	&82.5	&84.7 \\
                &$\surd$ 	&89.2	&83.5	&86.2 \\
\hline
\multirow{2}{*}{FAST}   &$\times$	&89.9	&83.2	&86.4 \\
                        &$\surd$ 	&92.8	&84.0	&88.2 \\
\hline
\end{tabular}
}
\vspace{-1mm}
\end{table}

\begin{table}[t]
\caption{Performance evaluation on different versions of FSNet on Total-Text. Best scores are highlighted in bold.
\vspace{-2mm}
}
\label{table:version result}
\setlength{\tabcolsep}{1.6mm}
\centerline{
\begin{tabular}{lccccc}
\hline
Version  &Paras(M)   &Prec.(\%)  &Recall(\%) &F1(\%)  &FPS\\
\hline
Ver. 1 &{\bf 29.3}  &93.1  &{\bf 87.0}	&{\bf 89.9} & 15.2 \\ 
Ver. 2 &15.1  &90.3  &85.5	&87.9 & 21.0 \\
Ver. 3 &27.2  &92.4  &86.1	&89.2 & {\bf 21.1} \\
Ver. 4 &28.6  &{\bf 93.6}  &84.6	&88.9 & 16.7  \\
\hline
\end{tabular}
}
\vspace{-4mm}
\end{table}

\subsection{Datasets}


We use Total-Text~\cite{totaltext} and ICDAR-ArT~\cite{icdar2019}, to evaluate the challenging cases, where the scene texts come with arbitrary shapes. Additionally, we employ MSRA-TD500~\cite{msra-td500} as a validation dataset for multi-directional scene texts.

\noindent{\bf Total-Text} is a challenging scene text detection benchmark dataset containing arbitrarily shaped text, horizontal, multi-directional, and curved text lines. The dataset comprises 1,555 images, with 1,255 images allocated for training and 300 images for testing. All text instances are annotated with word-level granularity.

\noindent{\bf ICDAR-ArT} is a large curved text dataset containing 5,603 training images and 4,563 testing images. It includes numerous text instances that are horizontal, multi-directional, or curved. We employ both the Total-Text and ArT datasets to assess the detection capability of MixNet on curved texts.

\noindent{\bf MSRA-TD500} comprises 500 training images and 200 test images, featuring multilingual, multi-oriented long texts (curved texts are not included). We use this dataset to evaluate the detection ability on multi-oriented texts. We include an additional 400 images from HUST-TR400~\cite{hust-tr400} for training in this evaluation.

\noindent{\bf SynthText}~\cite{synthtext} is a synthetic dataset containing 800K text images. All images are generated from 8K background images and 8 million synthetic word instances with word-level and character-level annotations. This dataset is used to pre-train our model.

\subsection{Implementation Details}

We used FSNet as the backbone and pre-trained it for five epochs on SynthText in our experiments. During pre-training, we utilized Adam optimizer and fixed the learning rate at 0.001. In the fine-tuning stage, we trained the model for 600 epochs on Total-Text, ArT, and other datasets, with an initial learning rate of 0.001 that decayed to 0.9 after every 50 epochs. The input image size was set to $640 \times 640$, and we employed data augmentation techniques such as random rotation $(-30^o \sim 30^o)$, random cropping, random flipping, and color jittering. The code was implemented using Python 3 and the PyTorch 1.10.2 framework. The training was carried out on an RTX-3090 GPU with 24G memory.

\begin{table}[t]
\caption{Ablation study of MixNet on Total-Text.
\vspace{-2mm}
}
\label{table:ablation:TotalText}
\centerline{
\setlength{\tabcolsep}{1.2mm}
\footnotesize
\begin{tabular}{ccccccc}
\hline
Pretrain  &FSNet &Center line   &Prec.(\%)  &Recall(\%) &F1(\%)  &FPS\\
\hline
$\times$ &$\times$ &$\times$ &91.4 &80.5 &85.7 & {\bf 21.7} \\
$\times$ &$\surd$ &$\times$ &91.1 &83.7 &87.3 &15.7 \\
$\times$ &$\surd$ &$\surd$ &{\bf 92.0} &{\bf 84.1} &{\bf 87.8} & 15.2 \\
\hline
$\surd$ &$\times$ &$\times$ &91.8 &84.0 &87.7 & {\bf 21.7} \\
$\surd$ &$\surd$ &$\times$ &{\bf 93.1} &87.0 &89.9 &15.7 \\
$\surd$ &$\surd$ &$\surd$ &93.0 &{\bf 88.0} & {\bf 90.5} &15.2 \\
\hline
\end{tabular}
}
\vspace{-1mm}
\end{table}

\begin{table}[t]
\caption{Ablations study on the effects of using other backbones on Total-Text.
\vspace{-2mm}
}
\label{table:ablation:backbones}
\centerline{
\setlength{\tabcolsep}{1.3mm}
\begin{tabular}{lccccc}
\hline
Backbone  &Paras(M)   &Prec.(\%)  &Recall(\%) &F1(\%)  &FPS\\
\hline  
ResNet50 &25.5  &88.2 &82.9 &85.5 &{\bf 21.7} \\
ResNet101 &44.5 &91.3 &82.0 &86.4 &17.8 \\
\hline
HRNet\_w18 &9.6  &87.9	 &74.5	&80.7 & 14.5 \\
HRNet\_w32 &28.5  &90.9	 &82.3	&86.4 & 13.7 \\
HRNet\_w48 &65.8  &91.3	 &82.7	&86.8 & 12.9 \\
\hline
FSNet\_S &27.2  &{\bf 91.9} &{\bf 84.2} &{\bf 87.9} &21.1 \\
FSNet\_M &29.3 &90.7 &84.0 &87.3 &15.7 \\
\hline
\end{tabular}
}
\vspace{-4mm}
\end{table}

\subsection{Accuracy Improvement of FSNet}

\begin{table*}[t]
\caption{Comparison between MixNet and other methods on the MSRA-TD500 dataset. 
Best scores are highlighted in bold.
\vspace{-2mm}
}
\label{table:MSRA}
\centerline{
\begin{tabular}{llcccc}
\hline
Method  &Backbone   &Prec.(\%)  &Recall(\%) &F1(\%) \\
\hline
CRAFT~\cite{craft} &ResNet50 &88.2	&78.2	&82.9 \\
PAN~\cite{PAN}	&ResNet18 &84.4	&83.8	&84.1 \\
DB~\cite{DB} &ResNet50-DCN	&91.5	&79.2	&84.9 \\
DB++~\cite{DB++} &ResNet50-DCN &91.5	&83.3	&87.2 \\
FAST~\cite{fast} &NAS-based	&92.1	&83.0	&87.3 \\
\hline
MixNet (Ours) &FSNet\_S &{\bf 92.3} &84.6 &88.3 \\
MixNet (Ours) &FSNet\_M	&90.7	&{\bf 88.1}	&{\bf 89.4} \\ 
\hline
\end{tabular}
}
\vspace{-1mm}
\end{table*}

\begin{table*}[t]
\caption{Comparison between MixNet and other methods on the Total-Text dataset.
Best scores are highlighted in bold.
\vspace{-2mm}
}
\label{table:TotalText}
\centerline{
\begin{tabular}{llcccc}
\hline
Method  &Backbone   & Prec.(\%)  & Recall(\%) & F1(\%)  & FPS \\
\hline
DB~\cite{DB} &ResNet50-DCN &87.1   &82.5   &84.7    &32.0   \\
PAN~\cite{PAN}  &ResNet18    &89.3   &81.0   &85.0   &{\bf 39.6}\\
CRAFT~\cite{craft} &ResNet50    &87.6   &79.9   &83.6   &8.6\\
I3CL~\cite{i3cl} &ResNet50  &89.8   &84.2   &86.9   &7.6  \\
FSG~\cite{fsg}  &ResNet50   &90.7   &85.7   &88.1   &13.1 \\
TextFuseNet~\cite{textfusenet} &ResNet101     &89.0   &85.8   &87.5   &3.7\\
TESTR~\cite{testr}  &ResNet50   &92.8   &83.7   &88.0    &5.3\\
TextBPN++~\cite{TextBPN++} &ResNet50   &91.8   &85.3   &88.5   &13.1\\
DPText-DETR~\cite{dptext}  &ResNet50   &91.8   &86.4   &89.0   &17.0\\
\hline
MixNet (Ours) &FSNet\_S &92.4 &86.1 &89.2 &21.1 \\
MixNet (Ours) &FSNet\_M &{\bf 93.0} &{\bf 88.1} &{\bf 90.5} & 15.2 \\
\hline
\end{tabular}
}
\vspace{-4mm}
\end{table*}

\begin{table}[t]
\caption{Performance comparisons between MixNet and others on ICDAR-ArT. Best scores are highlighted in bold.
\vspace{-2mm}
}
\label{table:ArT:result}
\centerline{
\renewcommand{\tabcolsep}{1mm}
\begin{tabular}{llccc}
\hline
Method  &Backbone   & Prec.(\%)  & Recall(\%) & F1(\%)\\
\hline
CRAFT &ResNet50  &77.2   &68.9   &72.9\\
TextBPN++ &ResNet50  &81.1   &71.1   &75.8\\
I3CL &ResNet50  &80.6   &75.4   &77.9\\
DPText-DETR  &ResNet50   &83.0   &73.3   &78.1\\
TextFuseNet &ResNet101   &{\bf 85.4}   &72.8   &78.6 \\
\hline
MixNet (Ours) &FSNet\_S &82.3 &75.0 &78.5 \\
MixNet (Ours) &FSNet\_M &83.0 &{\bf 76.7} & {\bf 79.7} \\
\hline
\end{tabular}
}
\vspace{-5mm}
\end{table}

Table~\ref{table:noise} provides empirical validation of the advancements achieved by the integration of high-resolution features in detecting small texts. In this table, text instances are categorized based on their sizes (Small, Medium, Large), and the detection performance is evaluated using Precision (Prec.), Recall, and F1-score. As evident in Table~\ref{table:noise}, FSNet greatly improves the success rates for detecting small and medium texts within the noise-free dataset.
We next report experiments assessing FSNet's noise resistance capabilities. We used impulse noise to simulate noise and used ResNet50 as the control group. Initially, ResNet50 and FSNet models were individually trained using a noise-free training dataset. Subsequently, noise points were introduced to the test data with probabilities of 5\% and 10\%, generating two noise-affected datasets. Finally, the trained models were used to detect text instances on the test data without noise, with 5\% noise, and with 10\% noise.
As in Table~\ref{table:noise}, within the noise-free dataset, FSNet exhibited a 2.1\% increase in F1-score compared to ResNet50. This improvement is further in the presence of noise, with enhancements of 4.5\% and 6.0\% for the two noise ratios, respectively. Although both models experienced a significant decline in performance with increasing noise levels, FSNet demonstrated superior noise resistance in noisy environments. This robustness is attributed to FSNet's architectural exchange of low-resolution and high-resolution features, which mitigates noise impact as features undergo convolution and pooling operations.

We next evaluate FSNet's capabilities by replacing the backbones and FPN architectures of existing models such as differentiable binarization (DB)~\cite{DB}, pixel aggregation network (PAN)~\cite{PAN}, and Faster Arbitrarily-Shaped Text (FAST) Detector~\cite{fast} with FSNet. We fixed the output channel of FSNet to be 256 for each scale, to generate image features across four scales. We apply $1\times1$ convolution to match the channel numbers with those of the original architecture. The training details adhered to the original methods protocol.
As shown in Table~\ref{table:differentMethods}, this replacement yielded a 1.5\% increase in F1-score for both the DB and PAN architectures and a substantial 1.8\% increase for FAST. This result shows that FSNet can successfully improve multiple existing methods.

\subsection{Ablation Study}


\begin{figure}[t]
\centerline{
  \includegraphics[width=\linewidth]{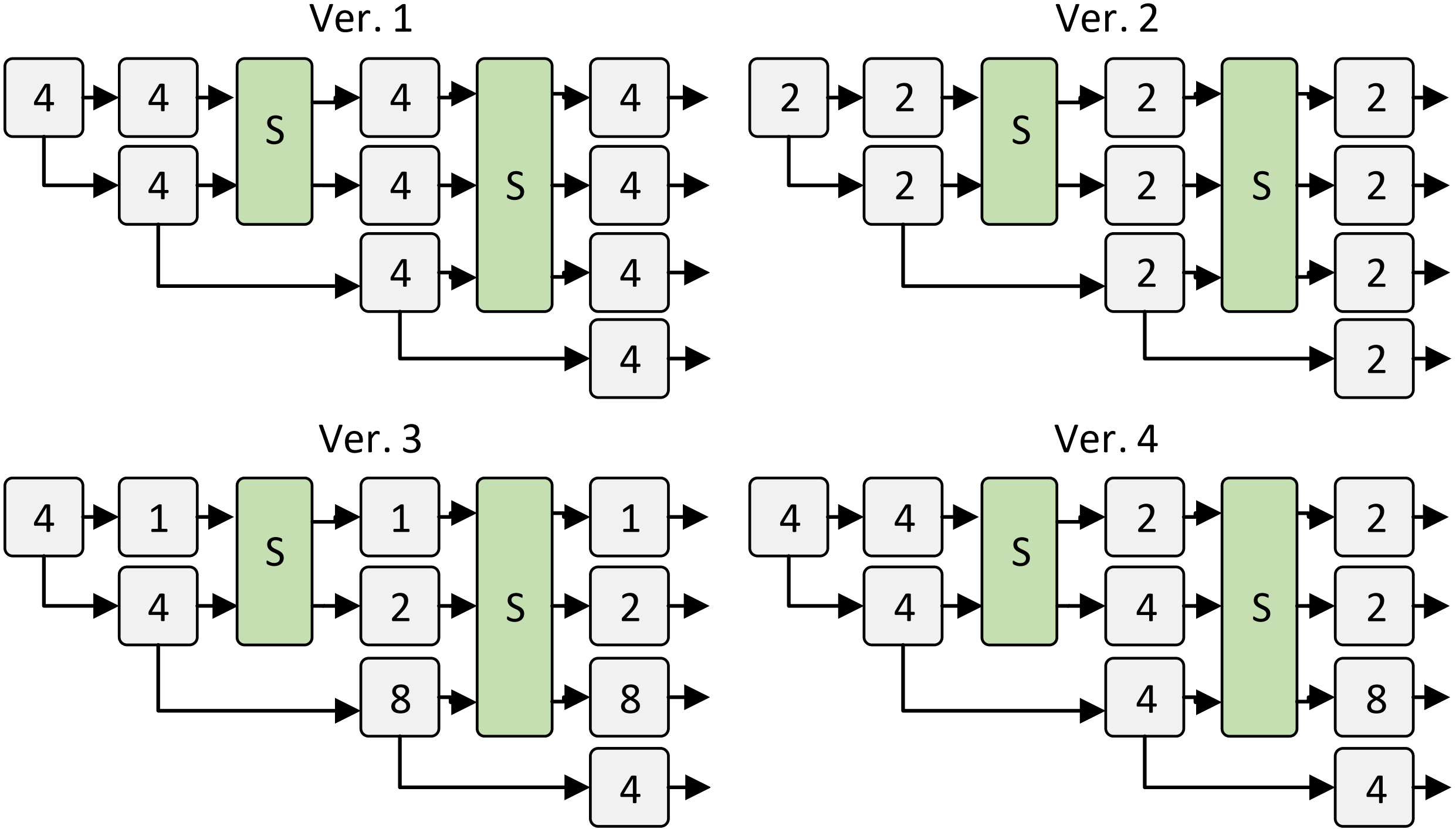}
  \vspace{-2mm}
}
\caption{The stacking numbers of each convolution block in four different versions of FSNet (note that the numbers within each square box denoting the stacking depth varies).}
\label{fig:FSNet:Vers}
\vspace{-.2in}
\end{figure}

\begin{figure*}[t]
\centerline{
  \includegraphics[width=\linewidth]{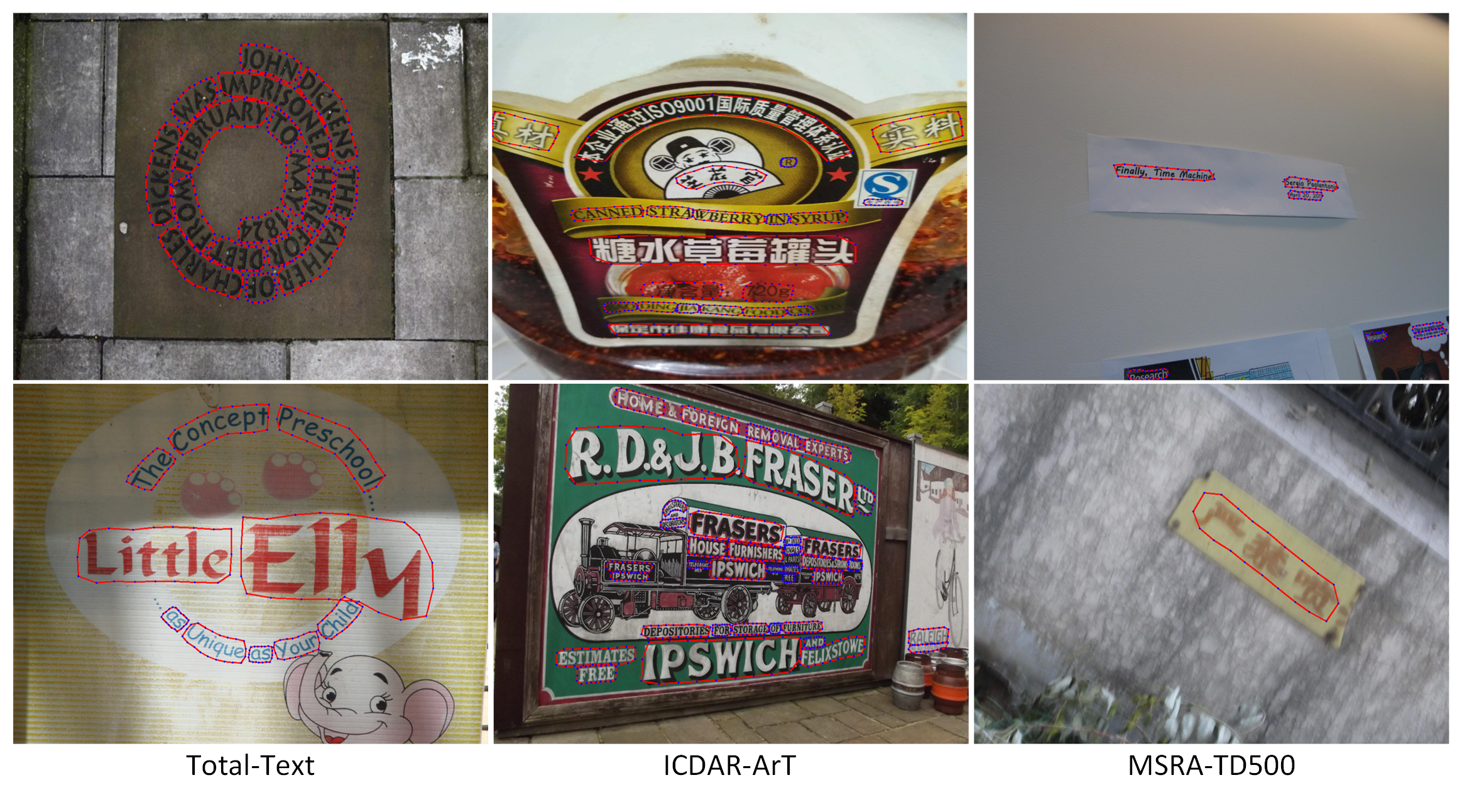}
\vspace{-6mm}
}
\caption{MixNet is capable of detecting small text from the scene, regardless of its orientation, shapes, and positions. 
}
\label{fig:demo}
\vspace{-4mm}
\end{figure*}

The stacking depth of each convolution block within FSNet was initially set to 4, denoted as Version 1. Despite delivering commendable performance, the computation load stemming from multi-layer convolutions on high-resolution features placed a strain on inference time. Consequently, we made adjustments to the convolution count within each block. Illustrated in Fig.~\ref{fig:FSNet:Vers}, we devised four iterations of FSNet, each featuring a distinct convolution count. In Version 2, we streamlined the stacking depth of each convolution block to 2, resulting in a significant boost in computational speed and a reduction in parameter count. However, this modification led to a noticeable decrease in performance on the dataset. Consequently, we opt to discard this version. In Versions 3 and 4, we implemented a reduction in the number of high-resolution convolutional layers, balanced by an increase in low-resolution convolutional layers to compensate. Scores and derivation speeds for all versions are itemized in Table~\ref{table:version result}. Clearly, the reduction of high-resolution convolutional layers triggers a marked increase in inference time. In our pursuit of striking a balance between performance and inference speed, we designated the convolution count of Ver. 3 as the blueprint for a lighter FSNet version, denoted as FSNet\_S. Additionally, for optimal results, the convolution count of Ver. 1 was retained as the foundational configuration for FSNet, named FSNet\_M.


Next, we validate the benefits of our newly proposed designs. We use the baseline using ResNet50 backbone which incorporates only contour-sampled points in the Transformer module. Table~\ref{table:ablation:TotalText} shows the results. In comparison, the use of FSNet backbone improves the scene text detection F1-score by 2.1\%, and the incorporation of center line features in CTBlock further improves the F1-score by 0.5\%.


\subsection{Comparison with Other Backbone}

Since the architecture of FSNet is similar to that of HRNet, we compared the parameters of the two architectures and their performance on Total-Text. Furthermore, since ResNet is the architecture most commonly used in text detection methods, we also compared ResNet in the table. As Table~\ref{table:ablation:backbones} shows, FSNet has a clear performance gap. However, despite having similar parameters to ResNet50, the high-resolution convolution operation and the up-sampling and down-sampling of the shuffle layer lead to a decrease in calculation time.

\subsection{Multi-oriented Scene Text Detection Result}

We use the MSRA-TD500 dataset to test the detection ability of MixNet on multi-oriented text detection. Consistent with other methods~\cite{DB,DB++}, we also use 400 training images from HUST-TR400 dataset. Table~\ref{table:MSRA} shows the performance comparisons between our method and other methods. Our architecture achieves an F-measure of 89.4\% on MSRA-TD500 dataset. Compared to other methods, our architecture improves the F measure by 2.1\%. This experiment proves that our architecture outperforms other methods on multi-oriented and multi-language text.

\subsection{Arbitrary-shaped Scene Text Detection Results}

As mentioned previously, the Total-Text dataset and ArT dataset hold significant importance in the realm of arbitrary-shaped text detection. These datasets encompass a wide range of text orientations, including horizontal, multi-directional, and curved text lines. Hence, we employ these two datasets to validate the capabilities of our architecture in detecting arbitrary-shaped text.
Our results in the TotalText dataset are presented in the last column of Table~\ref{table:TotalText}. Our architecture attains a remarkable F1-score of 90.5\%, thereby establishing a new state-of-the-art in performance. Compared to the prevailing methods, our F1-measure shows an improvement of 1.5\%, while the recall increases by 1.7\%. Moving to the ArT dataset, our architecture achieves an F-measure of 79.7\%. The comparative analysis between our approach and other methods is presented in Table~\ref{table:ArT:result}. Compared to TextFuseNet, our method improves the F1 score by 1.1\% and recall by 3.9\%. These experimental results underscore the fact that our architecture attains cutting-edge performance in arbitrary-shaped text detection.
Despite the varying orientations of the text instances, our method demonstrates a consistent ability to effectively detect them. Fig.~\ref{fig:demo} provides examples of text detection achieved by our method on diverse datasets. It becomes evident that MixNet excels at detecting straight lines, curves, and even densely arranged lengthy texts.


\section{Conclusion}

In this paper, we introduce MixNet, an architecture consisting of FSNet and the Central Transformer Block. FSNet enhances image features and minimizes noise interference in high-resolution data by shuffling features between low-resolution and high-resolution layers. Moreover, the Central Transformer Block incorporates a point sampling on contours and center lines. This design helps the model learn and optimize text boundaries, leading to more precise text contours.
Our experimental results underscore the efficacy of MixNet, as it achieves state-of-the-art performance on benchmarks for arbitrarily shaped scene text detection. Looking ahead, our future research involves exploring more effective methods for conveying low- and high-resolution features and delving into streamlined but highly representative point sampling approaches.

\bibliography{aaai24}

\end{document}


\maketitle
\section{Introduction}

In the supplementary material, we introduce a series of additional experiments aimed at providing additional evidence of the efficacy of our method. 

The first experiment in this study is related to the impact of the point sampling parameter $N$, which is consistently set to 20 in all the primary experiments in this paper. Through this experiment, we systematically evaluated the effect of varying the values of $N$ on the detection performance. 

The second experiment includes HRNet as an additional benchmark to extend the comparison between FSNet and ResNet in both noise-free and noisy scenarios presented in the main paper. Notably, both FSNet and HRNet maintain high-resolution features in this experiment. Through this setup, we validate that our feature shuffle strategy yields more substantial improvements compared to HRNet's feature fusion approach. By showcasing the superiority of our proposed feature shuffle strategy in enhancing performance, even in comparison to established techniques like HRNet's feature fusion, we contribute to the broader understanding of effective feature handling in text detection methodologies.

The third experiment delves into the influence of the presence of the shuffle layer on the detection ability, elucidating the enhancements brought about by our feature shuffle approach. Collectively, these supplementary experiments contribute to a comprehensive understanding of our method's effectiveness and provide deeper insights into the specific facets of our proposed approach.

\section{Number of Point Samples}
In the context of our paper, each text instance is encoded utilizing $N=20$ sampling points and subsequently refined through the Central Transformer Block. In general, a larger number of sampling points contributes to a more precise representation of the contour fit. Nevertheless, an escalated number of sampling points results in a heightened computational burden on the Central Transformer Block, thereby impacting the frame-per-second (FPS) performance. As a consequence, a judicious selection of the sampling number $N$ becomes imperative, striking an optimal equilibrium between detection accuracy and inference speed.

\begin{table}[]
  \caption{Performance evaluation on different point sample number N.}
  \label{sample-table}
  \centering
  \begin{tabular}{c|cccc}
  \hline
    $N$ &Prec.(\%)  &Recall(\%) &F1(\%) &FPS\\
    \hline
    12  &90.2	&83.4	&86.6 &15.9\\
    16 	&90.6	&83.1	&86.6 &15.5\\
    20	&91.1	&83.7	&87.3 &15.2\\
    24 	&91.0	&84.0	&87.4 &15.0\\
    28	&91.3	&84.1	&87.5 &14.8\\
    \hline
    \end{tabular}
  \label{table:inferenceofN}
\end{table}

The experiment is performed to explore the impact of varying the number of sampling points ($N$) on both detection performance and FPS. The results are summarized in Table~\ref{table:inferenceofN} for the $N$ values of $12$, $16$, $20$, $24$, and $28$. Notably, due to the inherent complexity of representing curved text contours, the experiment starts from a minimum of $N=12$. As indicated in the table, as $N$ increases, the FPS gradually decreases. However, the performance of MixNet remains stable when $N\geq20$. To achieve an optimal balance between detection accuracy and inference speed, we adopted $N=20$ for all of our experiments.

\section{Further Comparison with HRNet in Noisy Environment}

We conducted a performance comparison between FSNet and ResNet50 under conditions of both non-noise and noisy environments. We adopt HRNet, which shares architectural similarities with FSNet, as an additional benchmark for evaluation. In particular, it is important to note that both HRNet and FSNet effectively retain high-resolution features. As illustrated in Table~\ref{table:noise_hrnet}, in the context of a quiet environment, both HRNet and FSNet exhibit an edge over ResNet50 in the detection of small text instances. Therefore, in terms of FSNet's performance with the shuffle layer, it achieves a 1.8\% higher F1-score than HRNet.

Subsequently, when noise is introduced at levels of 5\% and 10\%, it becomes clear that HRNet experiences a notable decline in its ability to detect small text instances. Additionally, the performance gap between HRNet and FSNet also widens in the presence of noise. This observation underscores the robustness of FSNet's architecture in handling noisy environments, particularly in contrast to HRNet and ResNet50.  Our feature shuffling mechanism can bring semantic features from deep layer to shallow layer, resulting in this robustness improvement. 

\begin{table*}[t]
\caption{Performance evaluation on both noise-free and noisy datasets. Note how FSNet demonstrates better noise-resistant performance over ResNet50 and HRNet.
\vspace{-3mm}
}
\label{table:noise_hrnet}
\centerline{
\begin{tabular}{lccccccc}
\hline
Backbone	&Impulse noise &Prec.(\%)  &Recall(\%) &F1(\%) &Small(\%) &Medium(\%) &Large(\%)\\
\hline
HRNet\_w32   & \multirow{3}{*}{0\%} &91.6 &84.8 &88.0 &62.5 &88.7 &95.5\\
ResNet50    & 	&91.8	&84.0	&87.7  &53.0 &88.0 &94.5\\
FSNet      &	&92.1	&87.7	&89.8 \textbf{(+1.8)} &68.8 &92.4 &97.6 \\
\hline
HRNet\_w32   & \multirow{3}{*}{5\%} &90.3 &64.6 &75.3 &32.0 &66.0 &85.8\\
ResNet50	&	&85.0	&68.3	&75.8  & 26.9 &68.3 &89.2\\
FSNet      &	&90.4	&72.1	&80.3 \textbf{(+4.5)} &42.3 &75.3 &92.0\\
\hline
HRNet\_w32   & \multirow{3}{*}{10\%} &86.3  &53.8 &66.3 &15.3 &53.0 &79.0\\
ResNet50	&	&83.4	&58.9	&69.1 &16.5 &57.2 &84.9\\
FSNet      &	&88.6	&65.2	&75.1 \textbf{(+6.0)}  &30.7 &63.8 &88.0\\
\hline
\end{tabular}
}
\end{table*}

\begin{table*}[t]
\caption{ablation study of shuffle layer}
\label{table:shuffle layer}
\centerline{
\begin{tabular}{lccccccc}
\hline
Backbone &shuffle layer &Prec.(\%)  &Recall(\%) &F1(\%) &Small(\%) &Medium(\%) &Large(\%)\\
\hline
FSNet    &$\times$  &90.2	&85.6	&87.8   &58.7   &92.1   &95.2\\
FSNet    &$\surd$  	&92.1	&87.7	&89.8   &68.8   &92.4   &97.6 \\
\hline
\end{tabular}
}
\end{table*}

\section{Ablation Study of Shuffle Layer in FSNet}

The aim here is to delve into the analysis of the shuffle layer within FSNet. We establish a baseline by removing the shuffle layer from FSNet, forming the basis of our experimental comparison. As depicted in Table~\ref{table:shuffle layer}, the inclusion of the shuffle layer leads to an impressive increase in F1-score by 2.0\%. Moreover, this augmentation notably enhances the detection performance for both small- and medium-text instances, especially in small ones. Our feature shuffling mechanism can not only improve robustness to noise by bringing semantic features from the deep layer to the shallow layer, but also improve accuracy in small-sized text detection by bringing detailed spatial features from the shall layer to the deep layer. The experimental results show the importance of the shuffle layer in our method.

\begin{table}[t]
\caption{Performance comparisons between MixNet and others on CTW-1500
\vspace{-2mm}
}
\label{table:ctw1500}
\centerline{
\renewcommand{\tabcolsep}{1mm}
\begin{tabular}{llccc}
\hline
Method  &Backbone   & Prec.(\%)  & Recall(\%) & F1(\%)\\
\hline
PAN &ResNet18  &86.4   &81.2   &83.7\\
TextFuseNet &ResNet50   &85.8   &85.0   &85.4 \\
TextBPN++ &ResNet50  &87.3   &83.8   &85.5\\
I3CL &ResNet50  &87.4   &84.5   &85.9\\
\hline
MixNet (Ours) &FSNet\_S &89.2 &82.9 &86.0 \\
MixNet (Ours) &FSNet\_M &90.3 &84.8 &87.5 \\
\hline
\end{tabular}
}
\end{table}

\begin{table}[t]
\caption{Performance comparison between MixNet and DPText-DETR on CTW-1500 when using DPText-DETR's score program
\vspace{-2mm}
}
\label{table:ctw1500_fakescoreprogram}
\centerline{
\renewcommand{\tabcolsep}{1mm}
\begin{tabular}{llccc}
\hline
Method  &Backbone   & Prec.(\%)  & Recall(\%) & F1(\%)\\
\hline
DPText-DETR  &ResNet50   &91.7   &86.2   &88.8\\
\hline
MixNet (Ours) &FSNet\_S &89.5 &87.0 &88.2 \\
MixNet (Ours) &FSNet\_M &91.4 &88.3 &89.8 \\
\hline
\end{tabular}
}
\vspace{-5mm}
\end{table}

\section{Evaluation Performance on CTW1500 dataset}
Our results in the CTW1500 dataset are presented in the last column of Table~\ref{table:ctw1500}. Our architecture achieves an F1 score of 87.5\%, establishing a new state-of-the-art in terms of performance. Compared to other methods, our F1-score improves by 1.6\%. On the other hand, we found that the score program of DPText-DETR is different from the official one. For comparison with DPText-DETR, in Table~\ref{table:ctw1500_fakescoreprogram} we show the results using the same score program as DPText-DETR. As can be seen in the last column of Table~\ref{table:ctw1500_fakescoreprogram}, our F1-socre has increased by 1.0\%, and the recall has increased by 2.1\%.

\section{Visualization Results and Case Studies}

\begin{figure*}[t]
\centerline{
  \includegraphics[width=\linewidth]{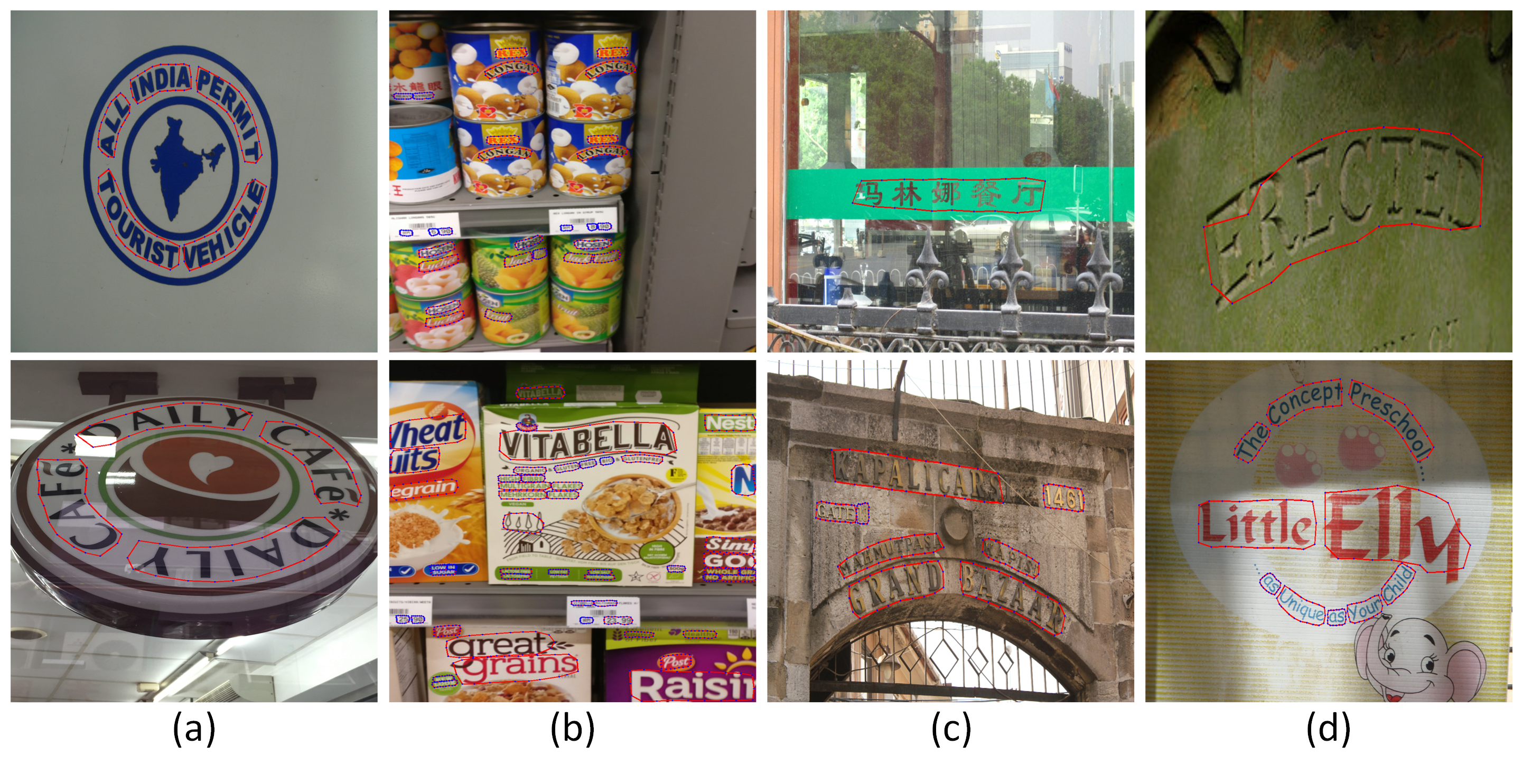}
}
\caption{Visualization results for various scene text detection examples. (a) corresponds to curved text, (b) depicts small-sized text, (c) illustrates a noisy background, and (d) showcases variations in lighting and shadows.}
\label{fig:GoodDetection}
\end{figure*}

In this section, we present a collection of visualization results to offer a more comprehensive understanding. Figure~\ref{fig:GoodDetection} showcases the detection outcomes across various text scenarios. Notably, MixNet exhibits proficient performance in scenarios involving (a) curved text, (b) small-sized text, (c) noisy backgrounds, and (d) variations in lighting and shadows.

\begin{figure*}[t]
\centerline{
  \includegraphics[width=\linewidth]{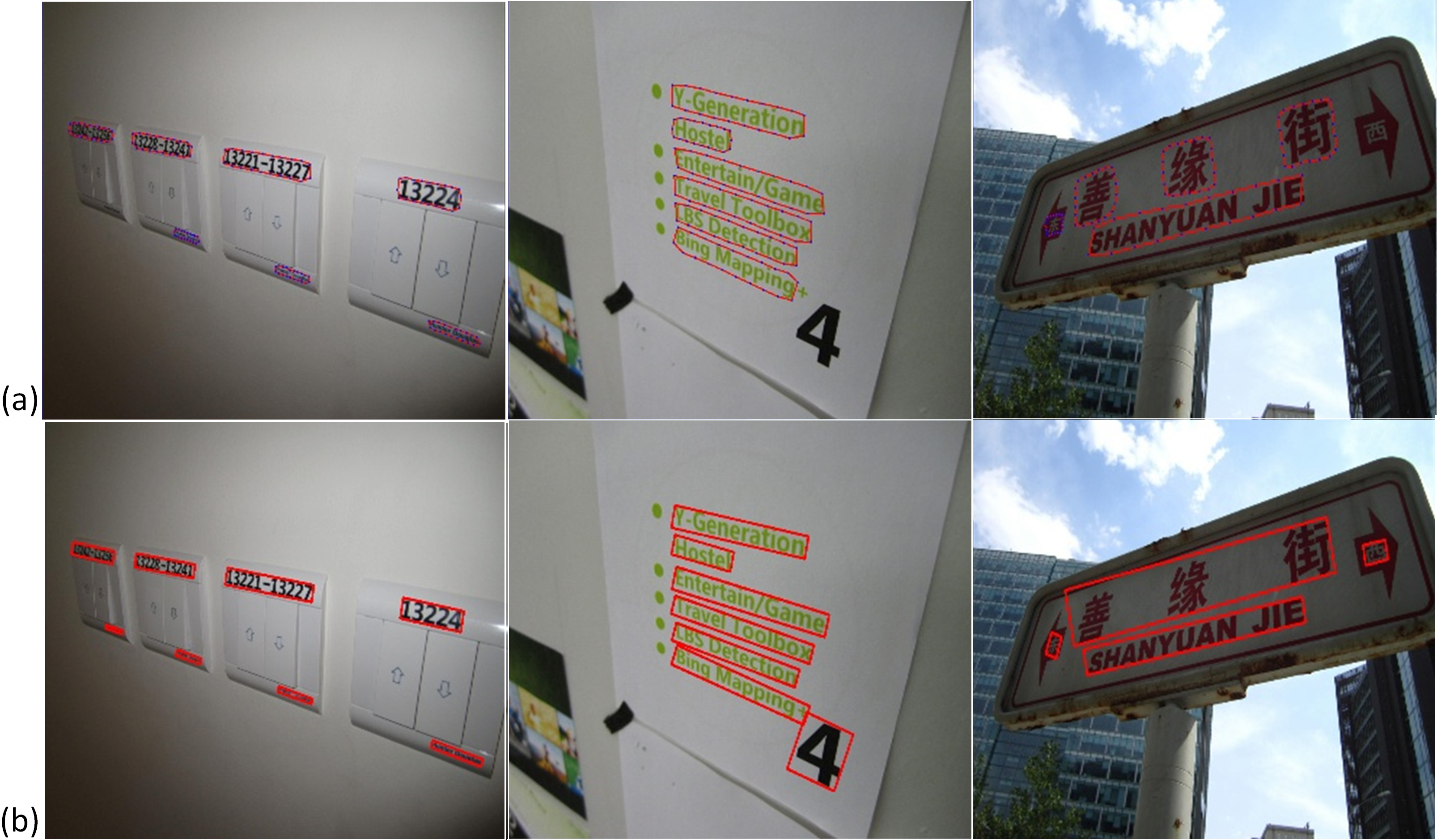}
}
\caption{Visualization results for a few cases where some scene texts are not 100\% detected. (a) corresponds the prediction of MixNet, (b) represent the ground-truth of images.}
\label{fig:faultcase}
\end{figure*}

Furthermore, Figure~\ref{fig:faultcase} illustrates instances where MixNet scene text detections are not 100\% correct. Progressing from left to right, three distinct issues are highlighted: (1) detection of extremely small text, (2) detection of single characters, and (3) incorrect segmentation. In the depiction, (a) represents the detection result produced by MixNet, while (b) illustrates the corresponding ground-truth. In the mentioned cases, although MixNet demonstrates satisfactory detection rates for small characters, consistent detection of extremely diminutive characters remains a challenge. Moreover, individual characters occasionally risk being overlooked, likely due to the bias in the training data towards instances with multiple characters, leading to an imbalance in feature representation for single characters and background elements. Lastly, the example at the right showcases a situation where the white space between text elements mirrors the background feature, causing MixNet to rely heavily on adjacent text features to identify key regions. Substantial white spaces exceeding a certain threshold can disrupt the connection of text features, potentially leading to erroneous detections.

